\documentclass[letterpaper, 10 pt, conference]{ieeeconf}  

\IEEEoverridecommandlockouts                              

\usepackage{graphicx} 
\usepackage{amsmath} 
\usepackage{amssymb}  

\usepackage{amsthm}
\usepackage{mathtools}
\usepackage{soul}
\usepackage{dsfont}
\usepackage[dvipsnames]{xcolor}
\usepackage{gensymb}

\usepackage{bm}
\theoremstyle{definition}

\newtheorem{remark}{Remark}

\usepackage{placeins}

\usepackage{hyperref}
\usepackage{algorithm}  
\usepackage{algpseudocode}[noend]

\usepackage{cite}
\usepackage{booktabs}

\graphicspath{{figures/}} 

\title{\LARGE \bf
BR-MPPI: Barrier-Rate Guided MPPI for Enforcing Multiple Inequality Constraints with Learned Signed Distance Fields
}

\author{Hardik Parwana$^{1, *}$, Taekyung Kim$^{1, *}$, Kehan Long$^{2}$, Bardh Hoxha$^{3}$, Hideki Okamoto$^{3}$, \\ Georgios Fainekos$^{3}$, and Dimitra Panagou$^{1}$  \\ 
\thanks{$^{*}$These authors contributed equally to this work}
\thanks{$^{1}$Robotics Department, University of Michigan, Ann Arbor, MI 48109, USA {\tt\small \{hardiksp, taekyung, dpanagou\}@umich.edu}}%
\thanks{$^{2}$Contextual Robotics Institute, University of California San Diego, La Jolla, CA 92093, USA. {\tt\small k3long@ucsd.edu}.}%
\thanks{$^{3}$Toyota Motor North America, Research \& Development, Ann Arbor, MI 48105, USA. {\tt\small <first\_name.last\_name>@toyota.com}}%
\thanks{This work was primarily conducted during the summer internships of Hardik Parwana, Taekyung Kim, and Kehan Long at Toyota Motor North America R \& D.}%
}

\newcommand{\reals}{\mathbb{R}}
\newcommand{\s}{\mathcal{S}}

\newcommand{\K}{\mathcal{K}}
\newcommand{\classK}{class-$\K$ }

\newcommand{\classKinf}{\mbox{class-$\K_\infty$}}

\newcommand{\eqn}[1]{\begin{align}#1\end{align}}

\begin{document}

\maketitle
\thispagestyle{empty}
\pagestyle{empty}

\begin{abstract}
Model Predictive Path Integral (MPPI) control provides a sampling-based framework for optimal control, while Control Barrier Functions (CBFs) provide a principled means of enforcing safety constraints. We introduce BR-MPPI, which integrates \textit{CBF-like} conditions into MPPI's control sampling procedure. CBFs impose inequality constraints that bound the rate of change of barrier functions using a \classK function of the barrier value. We instead impose the CBF condition as an \textit{equality constraint} using a parametric linear \classK function and augment the system state with its parameter. The parameter's time derivative serves as an additional control input optimized by MPPI. We further design a cost function that promotes parameter values consistent with Nagumo's condition at the safe-set boundary, thereby encouraging safety. The resulting multiple state- and control-dependent equality constraints pose a challenge for random control sampling. We address this through state transformations and control projections inspired by manifold path planning that map sampled controls onto the constraint manifold. We also incorporate learned signed distance fields to represent robot geometry and reduce computation time. Simulations demonstrate improved sample efficiency over vanilla MPPI and higher navigation success rates across five robot models compared with safety-oriented MPPI variants. Hardware experiments on a quadrotor further demonstrate the method's ability to navigate constrained environments near safe-set boundaries.
\href{https://www.taekyung.me/br-mppi}{\textcolor{red}{[Project Page]}}\footnote{Project page: \href{https://www.taekyung.me/br-mppi}{https://www.taekyung.me/br-mppi}} \href{https://github.com/tkkim-robot/br-mppi}{\textcolor{red}{[Code]}} \href{https://youtu.be/ApCJOnsDrwI}{\textcolor{red}{[Video]}} 
\end{abstract}

\section{INTRODUCTION}

Autonomous systems increasingly incorporate learning-based techniques into
real-time operation, from active learning for estimating unknown quantities
to offline-learned dynamics models and signed distance fields. Deploying
these systems in constrained environments requires real-time kinodynamic
planning that accounts for robot geometry and permits operation near
safe-set boundaries. Model Predictive Path Integral (MPPI)
control~\cite{williams2018information} and Control Barrier Function
(CBF)-based quadratic programs (QPs)~\cite{ames2016control} offer
complementary approaches to this problem. MPPI optimizes trajectories over
a finite horizon but does not inherently guarantee inequality constraint
satisfaction, whereas standard CBF-QP controllers enforce instantaneous
safety conditions but can be myopic and yield suboptimal trajectories.

Vanilla MPPI~\cite{williams2018information} solves unconstrained optimal
control problems by sampling control sequences, evaluating the resulting
state trajectories, and updating the nominal sequence through a
cost-weighted average of the sampled controls. Cost penalties provide a
soft treatment of inequality constraints, leaving MPPI susceptible to
violations, particularly with few samples~\cite{tao2022control}. Penalty
weights may also require extensive tuning to navigate close to constraint
boundaries without violating them. Among the extensions developed to
improve MPPI~\cite{gandhi2021robust,yin2022trajectory,trevisan2024biased,carius2022constrained},
we focus on CBF-based approaches to constraint satisfaction.

\begin{figure}
    \centering
    \includegraphics[width=0.7\linewidth]{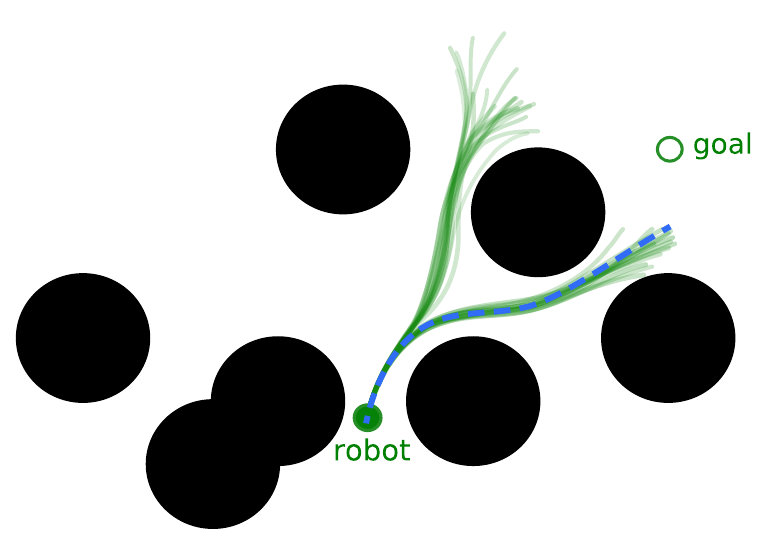}
    \caption{BR-MPPI navigation among obstacles (black). Green curves show
    sampled paths, and the blue curve shows the path generated by the
    updated control sequence. Planning in the augmented state space
    produces distinct groups of sampled paths on different sides of
    obstacles in the robot's original state space.}
    \label{fig:br_mppi_example_paths}
\end{figure}

A straightforward approach is to filter MPPI commands through a
CBF-QP~\cite{tao2022control}. Alternatively, the same work constructs trust
regions by optimizing the mean and covariance of the control sampling
distribution subject to CBF conditions. This requires a semidefinite
program (SDP) at each time step of every sampled trajectory, with fewer
samples used to mitigate the computational burden. Shield-MPPI~\cite{yin2023shield}
for deterministic systems and the method in~\cite{tsiotras2024chance}
for stochastic systems penalize CBF inequality violations and subsequently
apply gradient-based nonlinear optimization to enforce the constraints,
subject to convergence. GS-MPPI~\cite{rabiee2025guaranteed} instead applies
a closed-form CBF-based safety correction.

SC-MPPI~\cite{gandhi2023safe} incorporates barrier-state-based safe
feedback into sampling. The related formulation
in~\cite{gandhi2022safety} augments the robot state with barrier values
that diverge at constraint boundaries. These added states are functions
of the robot state: their evolution is determined by the robot
trajectory, rather than governed by additional control inputs.

We introduce Barrier-Rate Guided MPPI (BR-MPPI), which incorporates
\textit{CBF-like} conditions directly into sampling without the
inequality-constrained optimization used in~\cite{tao2022control}.
CBFs impose inequality constraints that bound the rate of change of
barrier functions using a \classK function of the barrier value.
For each constraint, we instead impose the CBF condition as an
\textit{equality constraint} using a parametric linear \classK function
and augment the system state with its parameter rather than the barrier
value. The parameter's time derivative serves as an additional control
input optimized by MPPI, allowing online adaptation of barrier evolution
without requiring singular barriers. We further design a cost function
that promotes parameter values consistent with Nagumo's condition at the
safe-set boundary, thereby encouraging safety.

The resulting multiple state- and control-dependent equality constraints
pose a challenge for random sampling, since sampled controls need not
satisfy them simultaneously. Inspired by manifold path planning, we
introduce state transformations and state-dependent control projections
that map sampled controls onto the constraint manifold. For control-affine
dynamics, a first-order approximation yields an analytical projection.
This formulation can induce multimodal trajectory distributions in the
robot's original state space, with sampled paths passing on different
sides of obstacles, as illustrated in
Fig.~\ref{fig:br_mppi_example_paths}.

To the best of our knowledge, this is the first work to consider:
1) MPPI with multiple state- and control-dependent equality constraints;
2) set invariance theory to promote constraint satisfaction in an augmented
state space accounting for barrier values and their rates of change; and
3) MPPI-based design or adaptation of CBF \classK function parameters.
Although parameter adaptation has been studied for
CBF-QPs~\cite{parwana2022recursive,parwana2022trust,xiao2021adaptive,zeng2021safety,ma2022learning},
it has not been studied for MPPI.

Simulation case studies demonstrate improved sample efficiency over
vanilla MPPI. In 100-trial benchmarks for each of five robot models,
BR-MPPI with analytical barriers achieves higher success rates than four
safety-oriented MPPI variants, with 87--100\% success and zero unsafe
trials. We also incorporate signed distance fields learned from robot
geometry, reducing mean command computation time by factors of
2.10--2.52 relative to analytical barriers, with model-dependent effects
on navigation outcomes. Quadrotor hardware experiments further demonstrate
navigation in constrained environments near safe-set boundaries.

\section{PRELIMINARIES}
\subsection{Notation}

The set of integers, real numbers, and positive real numbers are denoted as $\mathbb{Z}$, $\reals$, and $\reals^+$ respectively. The time derivative of $x$ is denoted by $\dot x$. Given $x\in \reals^n$, $||x||_Q\coloneqq\sqrt{x^T Q x}$ is called the weighted $Q$ norm for a positive definite matrix $Q$. $\langle a,b \rangle = a^Tb$ represents the inner product between $a,b \in \reals^n$. The boundary of a set $\mathcal{C}$ is denoted by $\partial C$. A continuous function $\alpha: [0,a)\rightarrow [0,\infty)$ is a \classK function if it is strictly increasing and $\alpha(0)=0$. Furthermore, if $a=\infty$ and $\lim_{r \rightarrow \infty} \alpha(r) = \infty$, then it is called \classKinf. Both $\frac{\partial }{\partial x}$ and $\nabla_x$ denote gradient and will be used interchangeably depending on the complexity of expressions for easy understanding.

\subsection{System Description}
Consider the following discrete-time dynamics with state $x_t\in \reals^n$, control input $u_{x_t} \in \reals^m$
\eqn{
x_{t+1} = x_t + F(x_t, u_{x_t})
\label{eq::dynamics_general}
}
where $F: \reals^n \times \reals^m\rightarrow \reals^n$ is the dynamics function. The following control-affine dynamics form, a special case of \eqref{eq::dynamics_general} will also be referred to later in Section \ref{section::simplify_affine_dynamics}
\eqn{
x_{t+1} = x_t + f(x_t) + g(x_t) u_{x_t}
\label{eq::affine_dynamics}
}
where $f:\reals^n \rightarrow \reals^n$ and $g:\reals^n \rightarrow \reals^{n \times m}$ are continuous functions. Let $\phi: \reals^n \rightarrow \mathbb{R}^p$ project the robot state $x_{t}$ to its position $\phi(x_{t}) \in \mathbb{R}^p$, we also denote the robot body as a set-valued function $\mathcal{B}(x_{t}) \subset \mathbb{R}^p$, and its surface as $\partial \mathcal{B}(x_t)$. 

The state $x_t$ is required to lie in safe sets $\s_i, i\in \{1,.., N\}$ that are defined to be 0-superlevel sets of continuously differentiable constraint functions $h_i: \reals^n \rightarrow \reals$
\begin{subequations}
    \begin{align}
        \s_i & \triangleq \{ x \in \reals^n : h_i(x) \geq 0 \} \\
        \partial \s_i & \triangleq \{ x\in \reals^n: h_i(x)=0 \}
    \end{align}
\end{subequations}

\subsection{Control Barrier Functions}
CBFs are a common tool for enforcing safety constraints in a controller. 
For a linear class-$\mathcal K$ function
$\nu_i(r)=\alpha_i r$ with $0<\alpha_i\leq 1$, the
following condition on the applied input preserves
$h_i(x_t)\geq 0$:
\eqn{
    h_i(x_t+F(x_t,u_{x_t}))-h_i(x_t)
    \geq -\alpha_i h_i(x_t).
    \label{eq::CBF_cond}
}
For multiple constraints, the same input must satisfy all
such inequalities simultaneously.
In practice, \eqref{eq::CBF_cond} is often imposed using a
constraint function $h_i$ as a candidate CBF, even when its
feasibility over the entire safe set is not guaranteed.
Intuitively, the CBF condition upper bounds the rate at which the $x_t$ can approach the safe set boundary: it allows $h(x_{t+1})<h(x_{t})$ when $h(x_t)>0$ but requires $h(x_{t+1})\geq h(x_t)$ when $h(x_t)=0$. The \textit{rate} is governed by $\alpha$. Since finding valid CBFs with desired performance characteristics (e.g., aggressive or conservative) is difficult $\alpha$ is usually tuned offline or online.

\subsection{Model Predictive Path Integral Control}

We briefly describe the MPPI algorithm. 
For a time horizon $H$, consider the state trajectory $\mathbf{x}=[x_t^T, ..., x_{t+H}^T]^T$, mean control input sequence $\mathbf{v} = [v_t^T, .., v_{t+H-1}^T]^T, v_\tau\in \mathbb{R}^m$, and Gaussian noise $\mathbf{w} = [w_t^T,..,w_{t+H-1}^T]^T$ where $w_\tau \sim N(0,\Sigma_w)$ and $\Sigma_w$ is the noise covariance, often chosen by the user. 
Let the disturbed control input sequence be $\mathbf{u}=[u_t,...,u_{t+H-1}]=\mathbf{v}+\bm{w}$.  
MPPI then solves the following problem
\begin{subequations}\label{eq::mppi_objective}
\begin{align}
\min_{\mathbf{v}}\quad J(\mathbf{v})
&=
\mathbb{E}\Biggl[
\Phi(x_{t+H})
\label{eq::mppi_objective_org}
\\[-1mm]
&\qquad
+
\sum_{\tau=t}^{t+H-1}
\left(
Q(x_\tau,u_\tau)
+
\frac{\lambda}{2}
v_\tau^\top\Sigma_w^{-1}v_\tau
\right)
\Biggr]
\nonumber
\\[-0.5mm]
\text{s.t.}\quad x_{\tau+1}
&=
x_\tau+F(x_\tau,v_\tau+w_\tau)
\\[-0.5mm]
w_\tau
&\sim
\mathcal{N}(0,\Sigma_w)
\label{eq::MPPI-disturbance}
\end{align}
\end{subequations}
Here, $Q:\reals^n\times\reals^m\to\reals$ is the stage
cost, $\Phi:\reals^n\to\reals$ is the terminal cost, and
$\lambda>0$ is the MPPI temperature.

MPPI uses sampling to solve \eqref{eq::mppi_objective}. Given a sample size $K$, we sample $K$ control input perturbation sequences $\bm{w}^k, k\in \{1,..,K\}$, where each sequence is of size $H$, that is $\bm{w}^k = [w_t^k, ..., w_{t+H-1}^k]$. The control input sequence is then computed as $\mathbf{u}^k=\mathbf{v}+\bm{w}^k$. For each sequence $k$, we compute the states $x^k_{\tau}, \tau\in [t,..,t+H]$, using dynamics \eqref{eq::dynamics_general}.

\eqn{
S_k
=
\Phi(x_{t+H}^k)
+
\sum_{\tau=t}^{t+H-1}
\left[
Q(x_\tau^k,u_\tau^k)
+\gamma v_\tau^T\Sigma_w^{-1}u_\tau^k
\right]
}
where $\gamma\in [0,\lambda]$. The weight $w_k$ of the $k^{th}$ trajectory is
\eqn{
w_k = \exp \left( -\frac{1}{\lambda} (S_k - \beta)  \right)
\label{eq::mppi_weight}
}
where $\beta=\min_{k\in\{1,..,K\}}S_k$. The optimal control sequence is then computed as
\eqn{
\mathbf{v}^+ = \sum_{k=1}^K w_k \mathbf{u}^k / \sum_{k=1}^{K}w_k.
\label{eq::mppi_final_control}
}

The cost functions are generally user-designed for specific objectives. We consider two such objectives: goal convergence and collision avoidance.

\subsection{Signed Distance Function}

Signed distance functions (SDFs) are widely used in robotics to accurately model environmental geometries \cite{luxin2019fiesta, oleynikova2017voxblox, Long_learningcbf_ral21}. Following recent developments \cite{long2024sensorbased_dro, li2024representing, long2024neural_df_continuum_robot}, we employ SDFs to model the robot's shape. This approach offers greater expressiveness for complex robot geometries and enables faster distance computations in environments represented by point-clouds. In particular, for a robot body $\mathcal{B}$, we define the robot's SDF $\varphi: \mathbb{R}^p \mapsto \mathbb{R}$:
\begin{equation}
\varphi(\mathbf{p}) := \begin{cases}
 -d(\mathbf{p},\partial \mathcal{B}), & \mathbf{p} \in \mathcal{B}, \\
 \phantom{-} d(\mathbf{p},\partial \mathcal{B}), &  \mathbf{p} \notin \mathcal{B},
 \end{cases}
\end{equation} 
where $d$ denotes the Euclidean distance between a point and a set, and $\mathbf{p} \in \mathbb{R}^p$ is a workspace point.

\section{METHODOLOGY}

We discuss our approach to training the robot SDF model.

\subsection{Training the Robot SDF Model}
\label{section::train_sdf}

To train the robot SDF, we prepare a dataset $\mathcal{D}$ from the robot's geometric description (e.g., URDF), containing samples $(\mathbf{p},d)$ of workspace points $\mathbf{p}\in\mathbb{R}^p$ and their signed distances $d$ to the robot surface $\partial\mathcal{B}$.

We train the SDF model $\hat{\varphi}(\mathbf{p}; \boldsymbol{\theta})$, parameterized by $\boldsymbol{\theta}$, using the loss
\begin{equation}
\ell(\boldsymbol{\theta}; \mathcal{D}) = \ell^D(\boldsymbol{\theta}; \mathcal{D}) + \lambda_E \ell^E(\boldsymbol{\theta}; \mathcal{D}),
\label{eq::sdf_loss}
\end{equation}
where $\ell^D$ and $\ell^E$ are the distance and Eikonal losses, respectively, and $\lambda_{E}$ balances them.
The distance loss $\ell^D$ measures the mean squared error between the predicted signed distances and the ground-truth signed distances 
$\ell^D(\boldsymbol{\theta}; \mathcal{D}) = \frac{1}{|\mathcal{D}|} \sum_{(\mathbf{p}, d) \in \mathcal{D}} \left( \hat{\varphi}( \mathbf{p}; \boldsymbol{\theta}) - d \right)^2 $.

The Eikonal loss $\ell^E$ encourages the SDF gradient to have unit norm like a true distance function
\begin{equation}
\ell^E(\boldsymbol{\theta}; \mathcal{D}) = \frac{1}{|\mathcal{D}|} \sum_{(\mathbf{p}, d) \in \mathcal{D}} \left( \left\| \nabla_{\mathbf{p}} \hat{\varphi}( \mathbf{p}; \boldsymbol{\theta}) \right\| - 1 \right)^2.
\end{equation}
By minimizing the combined loss function in \eqref{eq::sdf_loss}, we train the SDF model to accurately represent the robot's geometry.

\subsection{Safe MPPI Formulation}

In this work, instead of trying to impose the inequality constraint \eqref{eq::CBF_cond}, we impose the following equality constraint  
\eqn{
h_i(x_t + F(x_t,u_{x_t})) - h_i(x_t) = -\alpha_{i,t} h_i(x_t), \forall i\in \{1,..,N\}
\label{eq::cbf_equality}
}
and allow $\alpha_i$ to vary over time (hence the notation $\alpha_{i,t}$ in \eqref{eq::cbf_equality}). We achieve this by introducing a pseudo-parameter state $\tilde \alpha_{i,t}$ and additional control inputs $u_{\tilde \alpha_{i,t}}$ in the augmented system
\eqn{
\underbrace{\begin{bmatrix}
    x_{t+1} \\ \tilde \alpha_{t+1}
\end{bmatrix}}_{z_{t+1}} = \underbrace{\begin{bmatrix}
    x_{t} \\ \tilde \alpha_{t}
\end{bmatrix}}_{z_t} + \begin{bmatrix}
    F(x_{t}, u_{x_t}) \\ u_{\tilde \alpha_{t}}
\end{bmatrix}
\label{eq::augmented_dynamics}
}
where $z_t=[x_t^T ~ \tilde \alpha_t^T]^T\in \reals^{n+N}$ with $\tilde \alpha_t = [\tilde \alpha_{1,t},\ldots,\tilde \alpha_{N,t}]^T$ and $u_t=[u_{x_t}^T ~ u_{\tilde \alpha_t}^T]^T, u_{\tilde \alpha_t}=[ u_{\tilde \alpha_{1,t}}, .., u_{\tilde \alpha_{N,t}} ]^T$ denote the augmented state and control input. $\tilde \alpha_t$ will be used in Section \ref{section::design_projection_function} to compute the $\alpha_t$ for \eqref{eq::cbf_equality} at time $t$. Specifically, we will design parameter dynamics such that $\alpha_t=\tilde \alpha_{t+1}$. 

We allow $\alpha_t$ and $\tilde\alpha_t$ to vary in $\reals^N$.
For $h_i(x_t)>0$, negative $\alpha_{i,t}$ permits the barrier
value to increase, so the rate function is not restricted to
class-$\mathcal K$.
Section~\ref{section::mppi_cost} introduces a cost function
to promote constraint satisfaction.

\begin{remark}
    We introduce $\tilde \alpha_t$ rather than $\alpha_t$, as the new state variable because $\alpha_{t}$ and $u_t$ must satisfy \eqref{eq::cbf_equality} for $i\in \{1,..,N\}$ simultaneously. This compatibility cannot be enforced if $\alpha_t$ were a state variable that is guided by random MPPI sampled control inputs. Similarly, treating $\alpha_t$ as a control input would constrain the joint choice of $u_t$ and $\alpha_t$, precluding MPPI's random Gaussian sampling. We detail our procedure for ensuring compatibility below.
\end{remark}

\begin{remark}
    The notion of designing dynamics for parameter $\alpha$ is similar to existing works that design $\dot \alpha$ in discrete-time \cite{parwana2022recursive} and continuous-time \cite{parwana2022trust, xiao2021adaptive}.   
\end{remark}

Before presenting our final algorithm, we introduce another control space transformation. Let $u'_{x_t}\in \reals^{m}$ and $u'_{\tilde \alpha_t}\in \reals^{N}$ be pseudo-inputs that are randomly sampled by MPPI and $u_{x_t}$ and $u_{\tilde \alpha_t}$ are obtained from pseudo-inputs through a projection operation $\mathbb P: \reals^{n+N}\times \reals^m \times \reals^N\rightarrow \reals^{m} \times \reals^N$ introduced in the next section. MPPI therefore considers the dynamics
\begin{subequations}\label{eq::projected_augmented_dynamics}
\begin{align}
\begin{bmatrix}
x_{t+1}\\
\tilde{\alpha}_{t+1}
\end{bmatrix}
&=
\begin{bmatrix}
x_t\\
\tilde{\alpha}_t
\end{bmatrix}
+
\begin{bmatrix}
F(x_t,u_{x_t})\\
u_{\tilde{\alpha}_t}
\end{bmatrix},
\label{eq::projected_augmented_dynamics_state}
\\
\bigl(u_{x_t},u_{\tilde{\alpha}_t}\bigr)
&=
\mathbb{P}
\bigl(z_t,u'_{x_t},u'_{\tilde{\alpha}_t}\bigr).
\label{eq::projected_augmented_dynamics_input}
\end{align}
\end{subequations}
The projection $\mathbb P$ maps randomly sampled inputs onto the manifolds defined by \eqref{eq::cbf_equality}.

\subsection{Design of the Projection Function}
\label{section::design_projection_function}
The system of equations \eqref{eq::cbf_equality} might not admit a solution for arbitrary values of $u_{x_t}$ and $\alpha_t$. Therefore, we design $\mathbb P$ to ensure their compatibility for all $i\in \{1,..,N\}$.

Let $u'_{x_t}$ and $u'_{\tilde \alpha_t}$ be randomly sampled control inputs from MPPI. We project them onto the constraint manifolds as
\begin{subequations}\label{eq::u_alpha_proj}
\begin{align}
&
\begin{aligned}
\mathbb{P}(z_t,u'_{x_t},u'_{\tilde\alpha_t})
&=
\arg\min_{v,a}\;
\|v-u'_{x_t}\|_{Q_1}^2
\\[-1mm]
&\qquad\quad
+\|a-u'_{\tilde\alpha_t}\|_{Q_2}^2
\end{aligned}
\label{eq::u_alpha_proj_obj}
\\[-0.5mm]
&
\begin{aligned}
\text{s.t.}\quad
&h_i\!\left(x_t+F(x_t,v)\right)-h_i(x_t)
\\[-1mm]
&\qquad =
-(\tilde\alpha_{i,t}+a_i)h_i(x_t),
\quad i=1,\ldots,N .
\end{aligned}
\label{eq::u_alpha_proj_constraint}
\end{align}
\end{subequations}
Since $\tilde \alpha_{i,t}+a_i=\tilde \alpha_{i,t+1}$ the dynamics in \eqref{eq::projected_augmented_dynamics} ensure $\tilde \alpha_{t+1}=\alpha_t$.

\subsubsection{Simplification for Control Affine Dynamics}
\label{section::simplify_affine_dynamics}

The optimization \eqref{eq::u_alpha_proj} can be expensive for
general dynamics $F$. We therefore restrict our discussion to
the special case of control-affine dynamics in
\eqref{eq::affine_dynamics}, for which a first-order
approximation of \eqref{eq::u_alpha_proj} admits an analytical
solution. An extension to general dynamics is left for future
work. Under \eqref{eq::affine_dynamics}, linearizing $h$ about
$x_t$ gives
\eqn{
\begin{aligned}
&h\!\left(x_t+f(x_t)+g(x_t)u_{x_t}\right)-h(x_t)\\[-1mm]
&\qquad\approx
\left.\frac{\partial h}{\partial x}\right|_{x_t}
\left(f(x_t)+g(x_t)u_{x_t}\right).
\end{aligned}
}
making the constraints in \eqref{eq::u_alpha_proj} linear in $u_{t}$. The projection thus reduces to an equality-constrained weighted minimum norm problem:
\begin{equation}
\begin{aligned}
\mathbb{P}(z_t,u'_{x_t},u'_{\tilde{\alpha}_t})
={}&
\arg\min_{z}\quad
(z-z_{\textup{des}})^T W (z-z_{\textup{des}})
\\[-1mm]
&\text{s.t.}\quad
A_t z=b_t .
\end{aligned}
\label{eq::projection_lsq}
\end{equation}
where 
\eqn{
\begin{gathered}
z=
\begin{bmatrix}
v\\[-1mm]
a
\end{bmatrix},
\qquad
z_{\textup{des}}=
\begin{bmatrix}
u'_{x_t}\\[-1mm]
u'_{\tilde{\alpha}_t}
\end{bmatrix},
\qquad
W=\operatorname{diag}(Q_1,Q_2),
\\[1mm]
\begin{aligned}
(A_t)_{i,:}
&=
\begin{bmatrix}
\nabla h_i(x_t)^\top g(x_t)
&
h_i(x_t)e_i^\top
\end{bmatrix},
\\[-0.5mm]
(b_t)_i
&=
-\nabla h_i(x_t)^\top f(x_t)
-\tilde{\alpha}_{i,t}h_i(x_t).
\end{aligned}
\end{gathered}
\nonumber
}
Here, $e_i$ is the $i$-th standard basis vector of $\reals^N$,
and $Q_1\succ0$ and $Q_2\succ0$ are weighting matrices.

The solution to \eqref{eq::projection_lsq} is
\eqn{
  z = z_\textup{des} + W^{-1} A^T  ( A W^{-1}  A^T )^{-1}  (b - A z_\textup{des})
  \label{eq::ls_solution}
}

\begin{remark}
Sampling inputs subject to equality constraints is actively studied for path planning on manifolds \cite{kingston2018sampling}, e.g., for closed-chain systems with state constraints. 
Our projection is inspired by such works, as each equality constraint in \eqref{eq::cbf_equality} induces a manifold for the state-action pair. Consequently, our method inherits discretization errors associated with manifold constraints. While \eqref{eq::projection_lsq} performs well in practice, its first-order approximation can introduce errors for highly nonlinear $h$, potentially causing instability. 
One could recompute the tangent space of $h_i$ when this error exceeds a threshold, as in \cite{bordalba2020randomized} for kinodynamic RRT; we leave its application to real-time control for future work. Note that the dynamics \eqref{eq::dynamics_general} itself defines a manifold inherently followed by MPPI trajectories. However, to the best of our knowledge, this is the first work considering multiple state- and input-dependent equality constraints in MPPI.
\end{remark}

\begin{algorithm}[t] \footnotesize
\caption{BR-MPPI Controller}
\begin{algorithmic}[1]
    \Require $z_{t}$, $H$, $F$, $Q_c$ \Comment{current augmented state, time horizon, dynamics function, user-given convergence cost}
    \State Sample $w_\tau^k\sim\mathcal{N}(0,\Sigma_w)$,
    $k=1,\ldots,K$, $\tau=t,\ldots,t+H-1$
    \State $\mathbf{u}^k\leftarrow \mathbf{v} + \bm{w}^k$ 
    \For {$k=1$ to $K$} \Comment{Loop over samples; parallel in GPU}
        \State $z_t^k\leftarrow z_t$, $S_k\leftarrow0$ \Comment{Initialize Cost}
        \For {$\tau=t$ to $t+H-1$} \Comment{Loop over time horizon}
            \State $u'_{x_t}, u'_{\tilde \alpha_t} \leftarrow \bm{u}^k_\tau$ 
            \State Compute $u_{x_t}, u_{\tilde \alpha_t}$ using $\mathbb P$ in \eqref{eq::u_alpha_proj} or \eqref{eq::ls_solution}
            \State Compute next state $z_{\tau+1}$ using \eqref{eq::projected_augmented_dynamics}
            \State Compute $Q_h$ using \eqref{eq::constraint_cost}
            \State $S_k \!\leftarrow\! S_k + Q_c(x^k_{\tau}) + Q_h + \gamma (v_{\tau})^T \Sigma_w^{-1} (u^k_{\tau})$     
        \EndFor
        \State $S_k\leftarrow S_k+\Phi(x_{t+H}^k)$
        
    \EndFor
    \State $w_k \leftarrow $ \eqref{eq::mppi_weight} \Comment{Compute weights}
    \State $\mathbf{v}^+ \leftarrow $ \eqref{eq::mppi_final_control} \Comment{Computed final control trajectory} \\    
    \Return $\mathbf{v}^+$ 
\end{algorithmic}
\label{algo::mppi}
\end{algorithm}

\subsection{Design of MPPI Cost}
\label{section::mppi_cost}

We decompose the MPPI stage cost $Q$ into $Q_c$ and $Q_h$, which promote goal convergence and constraint satisfaction, respectively. $Q_c$ is user-designed; below, we define $Q_h$.

A necessary condition for constraint satisfaction is given by Nagumo's theorem \cite{nagumo1942lage, blanchini1999set} and states that whenever $h_i(x_t)=0$, $\dot h_i(x_t)\geq 0$, or in discrete time, $h(x_{t+1}) - h(x_t)=-\alpha_t h(x_t)\geq 0$. To account for discretization errors, we consider a buffer zone $D_i$ of length $d_i$ near the constraint boundary:
\eqn{
    D_i = \{ x ~|~ 0 \leq h_i(x)\leq d_i \}
}
Inside $D_i$, we promote $\alpha_{i,t}<0$ in \eqref{eq::cbf_equality} to increase $h_i$. The buffer length $d_i$ is a tuning parameter. Since $\alpha_{i,t}\in \reals$, the robot can explore different motions around obstacles while Nagumo's condition is promoted only near the boundary:
\eqn{
Q_h(z_t, z_{t+1}, u) = \sum_{i=1}^N \mathds{1}(x_t\in D_i \cup x_t \in \bar{\mathcal{S}}_i) \frac{\tilde \alpha_{i,t+1}}{h_i(x)}
\label{eq::constraint_cost}
}
The complete algorithm is summarized in Algorithm \ref{algo::mppi}.

\subsection{Discussion}

While applying MPPI to the augmented system \eqref{eq::augmented_dynamics} can be viewed as a state transformation, it has several important consequences. Here, we discuss its differences and improvements over standard MPPI and CBF approaches.

\subsubsection{Improvement of MPPI Through CBF}
First, we use the well-established theory of CBFs and set invariance to guide inequality constraint satisfaction. Our algorithm handles inequality constraints through two steps: 1) replacing each inequality constraint with a set of equality constraints parameterized by the additional state $\tilde \alpha$, and 2) using manifold planning techniques to satisfy these equality constraints. Their satisfaction is promoted through Nagumo's theorem near the constraint boundary.

Second, Fig.~\ref{fig:br_mppi_example_paths} shows sampled
trajectories passing on different sides of obstacles.
The state-dependent projection transforms Gaussian
pseudo-input samples before rollout, changing the resulting
trajectory distribution. Our method can still sample
trajectories inside obstacles, but avoids such solutions more
effectively in the reported examples.

\subsubsection{Improvement of CBF Through MPPI}

Finding valid CBFs remains an active research area, and candidate CBFs are commonly used in CBF-QP controllers to impose \eqref{eq::CBF_cond}. A candidate CBF may yield infeasible CBF-QPs for some states, while even a valid CBF may be overly conservative for a given \classK function. Several works therefore adapt the \classK function offline or online. Our algorithm can be viewed as adapting the \classK function online to improve performance while promoting safety. Moreover, although we use CBF-like conditions, constraint satisfaction is enforced through Nagumo's theorem only near the boundary, while $\tilde \alpha$ can vary in $\reals$ in the safe-set interior. Our method is therefore less restrictive than a fixed-\classK CBF. Further analysis of the relationship between our MPPI solutions and valid CBFs is left for future work.

\section{SIMULATION RESULTS}
\label{section::simulation}
Videos and code for all simulations and hardware experiments are accessible from our project page and paper attachments.

We use Algorithm \ref{algo::mppi} to navigate the robot in obstacle environments. We consider the following five dynamics models: single integrator (SI), unicycle (U), dynamic unicycle (DU), mobile arm (MA), and a planar quadrotor (PQ).
\eqn{
\textrm{SI: } & \dot p_x = u_1,~ \dot p_y = u_2 \label{eq::single_integrator} \\
\textrm{U: }& \dot p_x = v\cos\theta, ~\dot p_y = v\sin\theta, ~\dot \theta = u_1 \label{eq::unicycle} \\
\textrm{DU: }& \dot p_x = v\cos\theta, ~\dot p_y = v\sin\theta, ~\dot \theta = u_1,~ \dot v = u_2  \label{eq::extended_unicycle}\\
\textrm{MA: }& \dot p_x = u_1, ~\dot p_y = u_2, ~\dot \theta_{j,i} = u_{j,i}, j\in \{l, r\}, i\in \{1,2,3\} \label{eq::mobile_arm}  \\
\textrm{PQ: }& \ddot p_x = u_1, ~\ddot p_y = u_2, ~\ddot \theta = u_3 \label{eq::planer_quadrotor}
}
where $p_x,p_y\in\reals$ denote the planar position of the robot base, $\theta\in\reals$ its heading, and $v\in\reals$ its forward velocity. 
For the mobile arm, $\theta_{j,i}$ denotes the $i$th joint angle of arm $j$, where $j\in\{l,r\}$ identifies the left or right arm and $i\in\{1,2,3\}$. 
The variables $u_1$ and $u_2$ denote model-specific control inputs, corresponding to translational velocities, acceleration, angular velocity and acceleration as specified by the respective dynamics. For non-point robot footprints, we use the signed distance
representation described in Section~\ref{section::train_sdf}.

\begin{figure}
    \centering
    \includegraphics[width=1.0\linewidth]{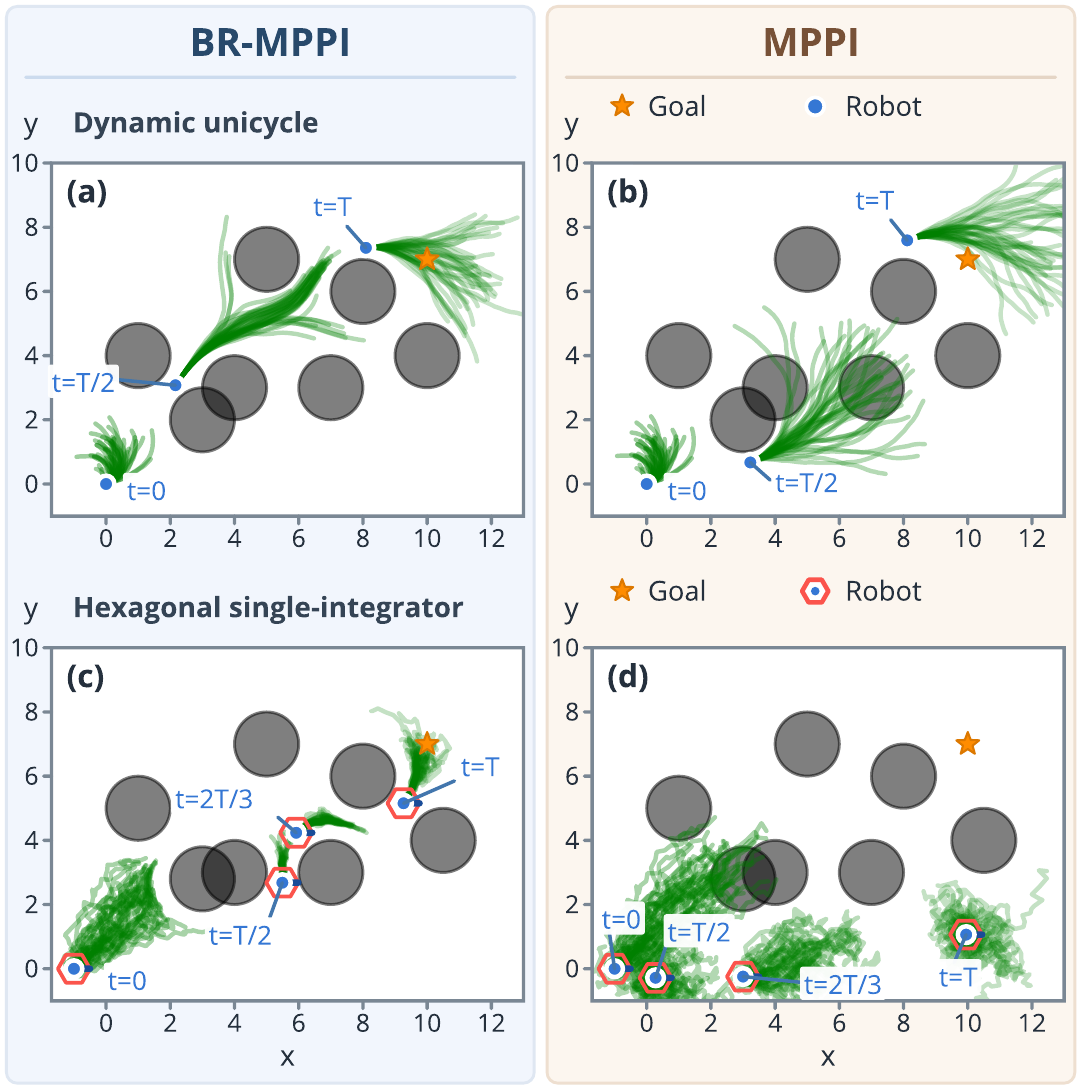}
    \setlength{\abovecaptionskip}{-7pt}
    \caption{(a),(b) Dynamic unicycle and (c), (d) Hexagonal single-integrator robots navigating an obstacle environment with BR-MPPI and MPPI. $T$ denotes the minimum of two values: the time required to reach within 2 meters of the goal (which varies across simulations) and the total simulation duration.}
    \label{fig:mppi_brmppi_comparison}
\end{figure}

\begin{figure*}[tp]
    \centering
    \includegraphics[width=\textwidth,height=0.86\textheight,keepaspectratio]{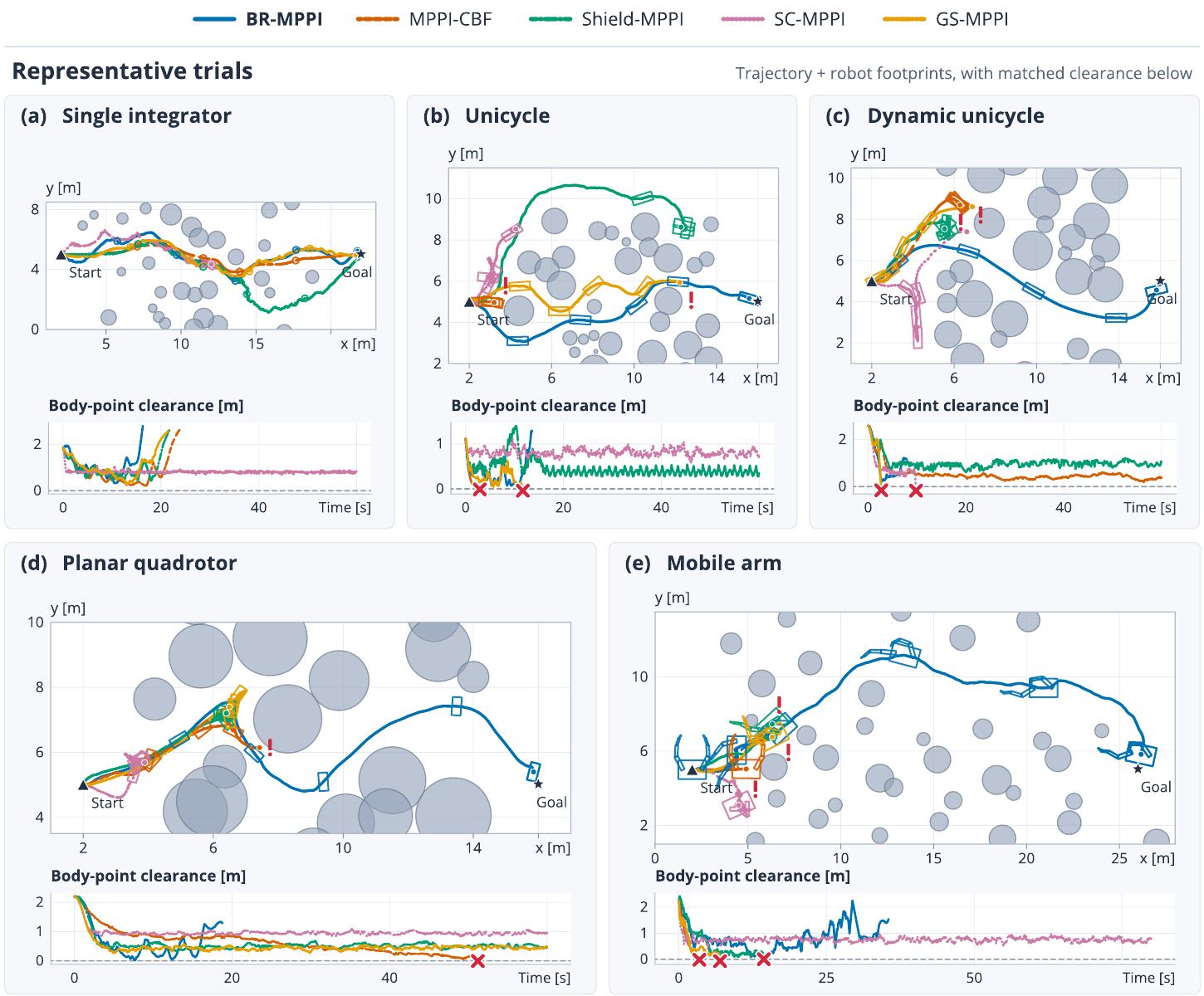}
    \setlength{\abovecaptionskip}{-7pt}
    \caption{
    Representative trials comparing BR-MPPI with MPPI-CBF,
    Shield-MPPI, SC-MPPI, and GS-MPPI across five robot models:
    (a) single integrator, (b) unicycle, (c) dynamic unicycle,
    (d) planar quadrotor, and (e) mobile arm.
    Each panel shows trajectories with robot footprints and the
    corresponding body-point clearance histories.
    Red exclamation marks in the trajectory plots and red crosses in the
    clearance plots indicate collisions.}
    \label{fig:five_robot_comparison}
\end{figure*}

\subsection{Comparative Study - Feasibility over Vanilla MPPI }
\label{section::vanilla_mppi_comparison}

We compare the proposed algorithm with vanilla MPPI to evaluate its ability to navigate robots with different shapes and dynamics without modifying the MPPI parameters, i.e., without fine-tuning MPPI for each new robot shape or dynamics model. We simulate both methods with $K=1000$, $\gamma=0.1$ and $\lambda=0.03$ in \eqref{eq::mppi_weight} but also normalize the costs to [0,1]. $Q_c(x)=\min(\|\phi(x)-p_d\|^2,100)$,
where $p_d\in\reals^p$ is the goal position and $Q_h$ is as defined in \eqref{eq::constraint_cost} for BR-MPPI and $Q_h = 1/\min_i(h_i)$ for MPPI. For second-order systems, we use the velocity-dependent
candidate $h_i=\delta_i+5\Delta t\,\dot\delta_i$, where
$\delta_i$ denotes signed obstacle clearance and $\Delta t$
is the simulation time step. Fig. \ref{fig:mppi_brmppi_comparison}(a), (b) shows snapshots of the paths generated by vanilla and proposed MPPI for a point-mass DU robot. The proposed algorithm successfully passes through narrow spaces, whereas vanilla MPPI takes a detour. Increasing the number of samples for vanilla MPPI to 20,000 results in paths similar to those obtained by our algorithm, demonstrating better sample efficiency of the proposed method in this case study. Next, we apply both algorithms to a hexagonal SI robot and visualize the resulting paths in Fig. \ref{fig:mppi_brmppi_comparison}(c), (d). Vanilla MPPI is overly conservative, causing the robot to remain near its initial position, whereas our algorithm successfully finds a path to the goal. Simulation videos for different robot dynamics are available on our website and the attachment.

\begin{figure*}[t]
    \centering
    \includegraphics[width=\textwidth]{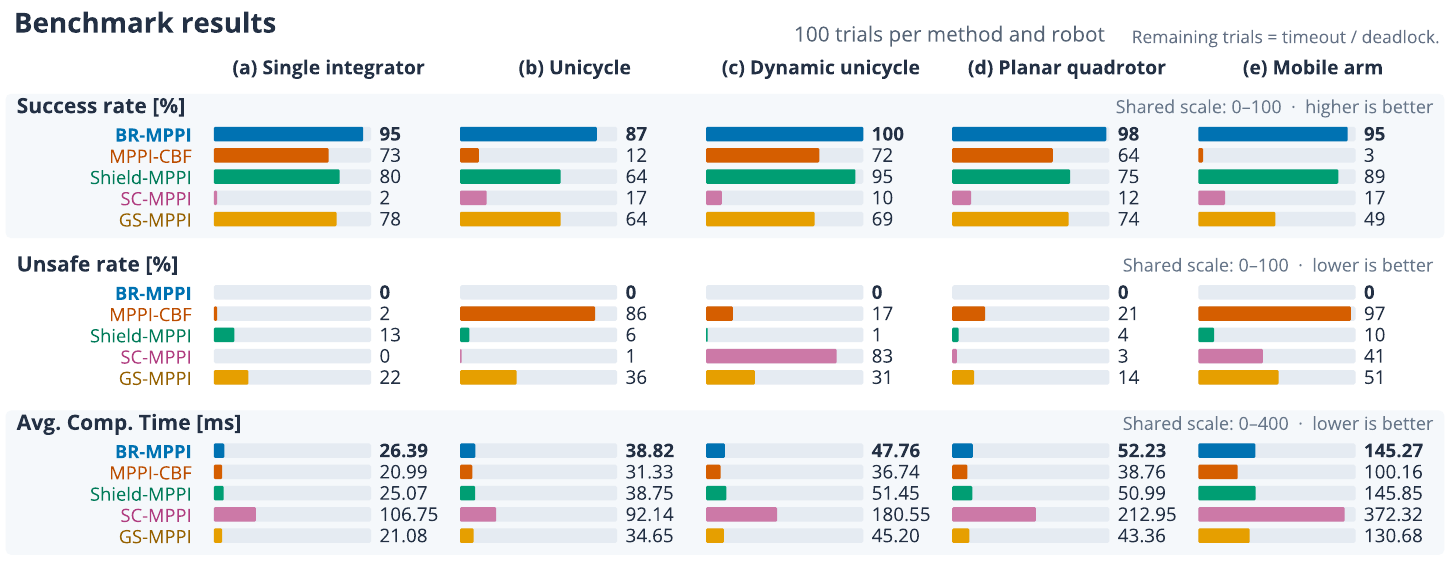}
    \setlength{\abovecaptionskip}{-7pt}
    \caption{
    Benchmark results over 100 trials per method and robot model.
    Columns (a)--(e) follow the robot-model order in
    Fig.~\ref{fig:five_robot_comparison}.
    Rows show success rate, unsafe rate, and mean command
    computation time.
    Trials that are neither successful nor unsafe terminate
    in timeout or deadlock.
    }
    \label{fig:five_robot_benchmark}
\end{figure*}

\begin{table*}[t]
    \centering
    \caption{Analytic versus optimized neural SDF barriers in
    BR-MPPI over 100 trials per robot model.
    R/C/T denotes reached/collision/timeout; deadlocks are included
    in timeouts. Speedup is the ratio of analytic to neural latency.}
    \label{table::neural_sdf_comparison}
    
    \small
    \renewcommand{\arraystretch}{1.08}
    \begin{tabular*}{\textwidth}{
        @{\extracolsep{\fill}}
        l c c c c c c c
        @{}
    }
        \toprule
        &
        \multicolumn{3}{c}{Analytic SDF}
        &
        \multicolumn{3}{c}{Optimized neural SDF}
        &
        \\
        \cmidrule(lr){2-4}
        \cmidrule(lr){5-7}
        
        Robot model
        & R / C / T
        & Success
        & \begin{tabular}[c]{@{}c@{}}Avg. Comp.\\Time (ms)\end{tabular}
        & R / C / T
        & Success
        & \begin{tabular}[c]{@{}c@{}}Avg. Comp.\\Time (ms)\end{tabular}
        & Speedup
        \\
        \midrule
        
        Unicycle
        & 87 / 0 / 13
        & 87\%
        & 38.82
        & 84 / 11 / 5
        & 84\%
        & 18.46
        & \textbf{2.10}$\times$
        \\
        
        Dynamic unicycle
        & 100 / 0 / 0
        & 100\%
        & 47.76
        & 100 / 0 / 0
        & 100\%
        & 18.97
        & \textbf{2.52}$\times$
        \\
        
        Planar quadrotor
        & 98 / 0 / 2
        & 98\%
        & 52.23
        & 94 / 0 / 6
        & 94\%
        & 23.48
        & \textbf{2.22}$\times$
        \\
        
        \bottomrule
    \end{tabular*}
\end{table*}

\subsection{Comparative Study - MPPI and CBF variants}
\label{section::mppi_variants_comparison}

We compare BR-MPPI against four safety-oriented MPPI baselines: MPPI-CBF, which filters the MPPI command through a CBF-QP; Shield-MPPI~\cite{yin2023shield}, which applies CBF-based shielding to the MPPI solution; SC-MPPI~\cite{gandhi2023safe}, which incorporates barrier-state-based safe feedback into sampling, while GS-MPPI~\cite{rabiee2025guaranteed} applies a closed-form CBF-based safety correction. We consider five dynamics---single integrator, unicycle, dynamic unicycle, planar quadrotor, and mobile arm---and tune all methods independently for each case study. Within each case study, the shared MPPI parameters, including the number of samples $K$ and horizon $H$, are fixed across methods, while method-specific hyperparameters are optimized independently using Optuna, with configurations selected primarily by task success. 

All methods share quadratic penalties on goal distance and control effort, together with a terminal goal-distance penalty. MPPI-CBF filters the selected action using a CBF-QP, while Shield-MPPI penalizes barrier violations and repairs the control sequence. SC-MPPI incorporates inverse-barrier states into its feedback optimization, whereas GS-MPPI applies a closed-form safety correction.
BR-MPPI, MPPI-CBF, and Shield-MPPI retain separate obstacle barriers. SC-MPPI combines them using a minimum when configured to do so, while GS-MPPI uses a smooth minimum. BR-MPPI additionally adapts a separate barrier-rate variable for each obstacle.

Figure~\ref{fig:five_robot_comparison} shows representative
trajectories and the corresponding body-point clearance histories
for the five methods across the five robot models. BR-MPPI reaches the goal without collision in all five
representative trials. In particular, panels~(c)--(e) illustrate
cases where the competing methods collide or fail to reach the goal.

In the 100-trial benchmarks shown in
Fig.~\ref{fig:five_robot_benchmark}, BR-MPPI achieves the
highest success rate
for every robot model (87--100\%) and is the only method with zero
unsafe trials across all five models. The remaining BR-MPPI trials
terminate in timeout or deadlock. Its mean command computation time
ranges from 26.39 to 145.27~ms, exceeding those of MPPI-CBF and
GS-MPPI but remaining below that of SC-MPPI for every model.

These violations do not contradict the baselines'
conditional safety guarantees. Distance-based candidate CBFs may be infeasible under bounded inputs; SC-MPPI relies on barrier-state feedback that is locally valid around the nominal trajectory; and GS-MPPI can lose its CBF guarantee when actuator saturation modifies the closed-form safe control.

\begin{figure*}[tp]
    \centering
    \includegraphics[
        width=\textwidth,
    ]{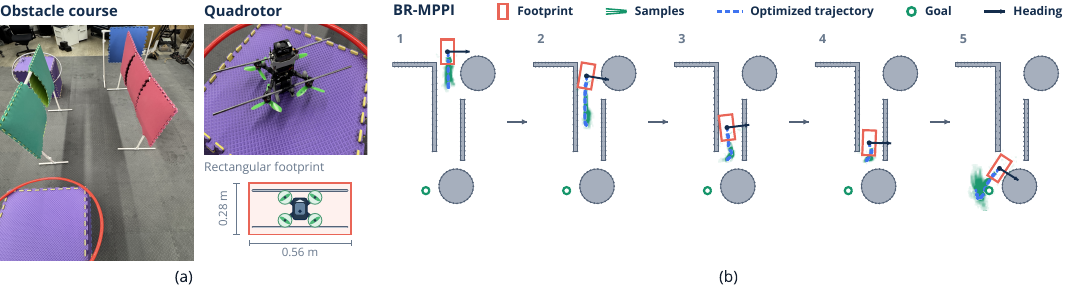}
    \setlength{\abovecaptionskip}{-7pt}
    \caption{(a) Experimental setup showing the obstacle course and our
    custom-assembled quadrotor. (b) Visualization of a hardware run with
    BR-MPPI. Video available on our project page.}
    \label{fig::quad_exp_setup_sim}
\end{figure*}

\subsection{Comparative Study: Exact vs Learned SDF}

We compare analytic and learned signed-distance barriers within
BR-MPPI over 100 trials for each of the three robot models in
Table~\ref{table::neural_sdf_comparison}. We retain the tuned BR-MPPI controller parameters from Section \ref{section::mppi_variants_comparison} and change only the barrier-function implementation from the analytical distance representation to its learned counterpart. The results are shown in Table \ref{table::neural_sdf_comparison}. The optimized neural
SDF reduces mean command latency by factors of
$2.10$--$2.52$, with model-dependent effects on navigation outcomes.
For the dynamic unicycle, both formulations achieve 100 successful
trials without collisions. For the planar quadrotor, the neural
formulation remains collision-free, while success decreases from
98\% to 94\%. For the unicycle, success decreases from 87\% to 84\%,
and 11 trials terminate in collision. While safety depends on the accuracy of the learned geometry, these results demonstrate the computational benefit of learned barriers in achieving real-time execution speeds.

\section{HARDWARE EXPERIMENTS}
\label{section::experiment}

We validate the proposed algorithm through hardware experiments on a
quadrotor navigating an environment with obstacles. The quadrotor is
fitted with metal straws to form a rectangular footprint of size
$0.28 \times 0.56$~m and follows the planar quadrotor dynamics in
\eqref{eq::planer_quadrotor}, with positions $p_x,p_y$ and heading
angle $\theta$. The system is implemented in ROS~2, with localization
provided by a Vicon motion-capture system. The experimental setup is
shown in Fig.~\ref{fig::quad_exp_setup_sim}(a), where the objective is
to reach the goal while avoiding obstacles.

Each obstacle is represented by a point cloud of 30 points.
Fig.~\ref{fig::quad_exp_setup_sim}(b) shows the corresponding
visualization of a hardware run with BR-MPPI. In the same environment,
vanilla MPPI is unable to enter the corridor, with its trajectories
omitted due to space constraints. We use the same parameters as in
Section~\ref{section::vanilla_mppi_comparison} and observe that, in
this highly constrained environment, vanilla MPPI fails to complete
the task, whereas BR-MPPI successfully reaches the goal. A video of
the hardware experiment is available on our project page and in the supplementary video.

By augmenting the state with $\tilde{\alpha}$ and optimizing its
discrete-time increments, we influence how the robot approaches or
moves away from obstacles. This capability is particularly important
in constrained environments, where feasible trajectories may require
operating close to constraint boundaries. A rigorous theoretical
characterization of how this augmented-state reformulation alters the
behavior of vanilla MPPI remains an important direction for future
work.

\section{CONCLUSION AND FUTURE WORK}

We presented BR-MPPI, which guides control sampling through CBF-like
\textit{equality constraints} by augmenting the state with barrier-rate
parameters and projecting sampled controls. Simulations demonstrate
improved sample efficiency over vanilla MPPI and higher success rates
than safety-oriented MPPI variants, while quadrotor experiments
demonstrate navigation near safe-set boundaries. Learned signed distance
fields reduce computation time, with model-dependent effects on
navigation outcomes. Future work includes incorporating higher-order
CBFs, establishing theoretical guarantees for constraint satisfaction
under actuation constraints, and extending hardware evaluations to
a broader range of robotic systems.

\bibliographystyle{IEEEtran}
\bibliography{icra.bib}

\end{document}